\pdfoutput=1

\documentclass[11pt]{article}

\usepackage{tabularx}
\usepackage{booktabs}
\usepackage{amsmath} 
\usepackage[]{acl}
\usepackage{graphicx}
\usepackage{algorithm}
\usepackage{cleveref}
\usepackage{xcolor}
\crefformat{section}{\S#2#1#3}
\crefformat{subsection}{\S#2#1#3}
\crefformat{subsubsection}{\S#2#1#3}
\crefrangeformat{section}{\S\S#3#1#4 to~#5#2#6}
\crefmultiformat{section}{\S\S#2#1#3}{ and~#2#1#3}{, #2#1#3}{ and~#2#1#3}

\Crefformat{figure}{#2Fig.~#1#3}
\Crefmultiformat{figure}{Figs.~#2#1#3}{ and~#2#1#3}{, #2#1#3}{ and~#2#1#3}
\Crefformat{table}{#2Tab.~#1#3}
\Crefmultiformat{table}{Tabs.~#2#1#3}{ and~#2#1#3}{, #2#1#3}{ and~#2#1#3}
\Crefformat{appendix}{#2Appx.~\S#1#3}
\crefformat{algorithm}{Alg.~#2#1#3}

\usepackage{algpseudocode}
\usepackage{booktabs}
\usepackage{times}
\usepackage{latexsym}
\usepackage{amssymb}
\usepackage{amsfonts}
\usepackage{xspace}
\usepackage{color,soul}
\usepackage[T1]{fontenc}

\usepackage[utf8]{inputenc}
\usepackage{multirow}
\usepackage{microtype}

\newcommand{\modelname}{\textsc{PLUTo}\xspace}
\newcommand{\modelmeaning}{\underline{\textsc{P}}lanning, \underline{\textsc{L}}earning, and \underline{\textsc{U}}nderstanding for \underline{\textsc{To}}ols\xspace}

\newcommand{\stitle}[1]{\vspace{1ex} \noindent{\bf #1.}}

\title{Planning and Editing What You Retrieve for Enhanced Tool Learning}

\author{
  Tenghao Huang\textsuperscript{1} \quad 
  Dongwon Jung\textsuperscript{1}  \quad 
  Muhao Chen\textsuperscript{2} \\
  \textsuperscript{1}University of Southern California \\
  \textsuperscript{2}University of California, Davis \\
  \texttt{\{tenghaoh, dongwonj\}@usc.edu;} \\
  \texttt{muhchen@ucdavis.edu;}
}

\begin{document}
\makeatletter\acl@finalcopytrue
\maketitle

\begin{abstract}
Recent advancements in integrating external tools with Large Language Models (LLMs) have opened new frontiers, with applications in mathematical reasoning, code generators, and smart assistants. However, existing methods, relying on simple one-time retrieval strategies, fall short on effectively and accurately shortlisting relevant tools. This paper introduces a novel \modelname (\modelmeaning) approach, encompassing ``Plan-and-Retrieve (P\&R)'' and  ``Edit-and-Ground (E\&G)'' paradigms. The P\&R paradigm consists of a neural retrieval module for shortlisting relevant tools and an LLM-based query planner that decomposes complex queries into actionable tasks, enhancing the effectiveness of tool utilization. The E\&G paradigm utilizes LLMs to enrich tool descriptions based on user scenarios, bridging the gap between user queries and tool functionalities. Experiment results demonstrate that these paradigms significantly improve the recall and NDCG in tool retrieval tasks, significantly surpassing current state-of-the-art models.

\end{abstract}
\begin{figure}[t]
    \centering
    \includegraphics[width=1.02\linewidth]{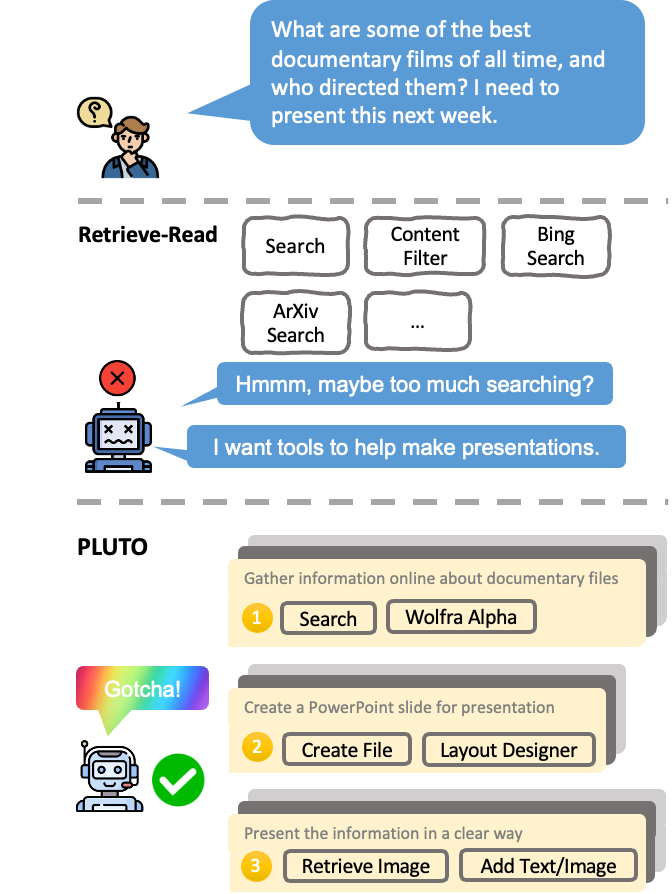}
    \caption{Comparison between conventional Retrieve-and-Read and \modelname paradigm. Unlike the conventional one-time Retrieve-and-Read paradigm that may lead to retrieving an ineffective set of tools, PLUTo efficiently parses a complex query and distills it into actionable sub-queries that facilitate accurate retrieval of appropriate tools.}     
    \label{fig:intro}
    \vspace{-1em}
\end{figure}


\section{Introduction}
The community has shown increasing interest in integrating external tools and interfaces with LLMs since tools often provide complementary functionalities in complex tasks such as dialogues \cite{bubeck2023sparks}, mathematical reasoning \cite{lu2022dynamic}, and code generation \cite{yadav2023exploring}. 
To realize tool augmentation, LLM systems typically employ a retriever mechanism to select relevant tools from a candidate pool and write function API calls based on the retrieved tools. 
The introduction of external tools also allows LLMs to address complicated user queries. \citealt{schick2023toolformer} show that LLMs, incorporating simple tools, achieve better performance on downstream tasks. \citealt{gupta2023visual} attempt to solve compositional visual tasks via image processing modules and language-instructed computer vision models. 
More recently, the integration of LLMs and tools empower LLMs, opening up new possibilities in areas like scientific discovery \cite{yang2023large}, automated efficiency, and smart assistant applications \cite{shu2022dialog2api}.

Nonetheless, emergent approaches for LLMs with tool integration present several distinct challenges. 
One primary concern is that current LLM agents still adopt simple retrieval-and-read strategies \cite{patil2023gorilla, qin2023toolllm}, lacking the dynamic adaptability required for addressing complex queries. As shown in \Cref{fig:intro}, 
 the conventional Retrieve-and-Read paradigm, solely relying heavily on similarity matching, falls short of retrieving diverse types of tools to address a complex user query. This limitation is further exacerbated by the semantic gap between user queries and tool descriptions. 
Particularly, user queries can be ambiguous and complex, often requiring a deep understanding of the user's intent and the context of the query \cite{kulkarni-etal-2023-label}. 
On the other hand, human-written tool descriptions can be abstract and lack essential details for deciding their utilities, leading to a mismatch between what the user needs and what the tool is perceived to offer. 
Additionally, current models tend to finetune on static tools, posing challenges to their robustness in the ever-evolving tool environment where new tools emerge and existing ones become obsolete \cite{api_life_cycle}. There is limited research on retrieval enhancement strategies in non-finetuned settings. These gaps highlight crucial areas for future research and development in LLM and tool integration.
\begin{figure*}[t]
    \centering
    \includegraphics[width=0.95\linewidth]{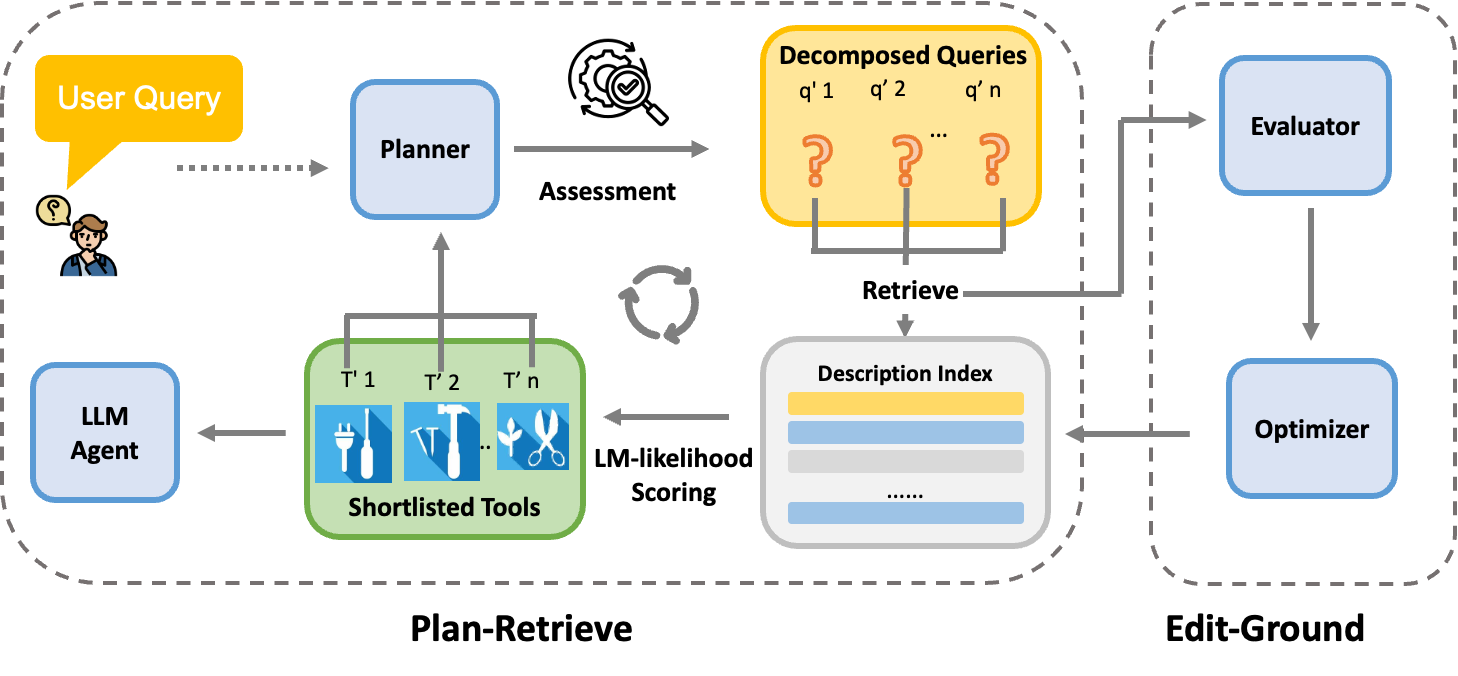}
    \caption{An overview of the \modelname approach.}     
    \label{fig:pipeline}
    \vspace{-1em}
\end{figure*}

In this paper, we leverage LLM's world knowledge and reasoning ability to augment the retrieval and utility of tools in response to complex user queries, by designing a novel framework \modelname~(\modelmeaning) \footnote{Code is available at \url{https://github.com/tenghaohuang/PLUTo} }. Our first contribution is the introduction of a novel \emph{Plan-and-Retrieve} for tool integration. While prior Retrieve-and-Read approaches only retrieve once at the beginning, our \emph{Plan-and-Retrieve} paradigm is designed to adaptively adjust its strategies based on the outcomes of its self-evaluations, ensuring a continuous refinement of the tool selection process. This paradigm is structured into two core modules. The first module, the retriever, leverages neural (dense) retrieval techniques \cite{karpukhin-etal-2020-dense} and LM-likelihood scoring mechanisms \cite{song2023llm} to efficiently shortlist relevant tools from a vast pool of candidates in response to a user query. This process ensures that the most pertinent tools are identified quickly, laying a foundation for more effective tool utilization. Inspired by recent advancements of adaptive retrieval-augmented generation (RAG; \citealt{jiang2023active, yoran2023making}), we design an LLM-based query planner that autoregressively decomposes complex user queries into manageable, task-oriented actions as the second module. Following the decompositions, the query planner selects the most suitable ones from the retrieved tools. It goes further by evaluating the effectiveness of selected tools and proposing the next action toward addressing the user query. This \emph{Plan-and-Retrieve} paradigm operates dynamically, embodying a sophisticated feedback loop that interlinks the retrieval of tools with subsequent refinement, evaluation, and planning stages. 

Our second contribution is the proposal of \emph{Edit-and-Ground} paradigm that utilizes user queries' rich contextual information and LLM’s extensive world knowledge for enriching descriptions of tool functionalities. Research has shown that informed tool documentations can enhance the interaction between LLMs and tools \cite{hsieh2023tool}. However, documenting tool functionalities at scale can be tedious for humans. \citealt{yang2023large} show LLMs can follow instructions and optimize real-world applications. Leveraging the optimization ability of the LLM, our tool-grounding agent optimizes 
under-informative tool descriptions by learning and abstracting information from tools' user scenarios. By editing tool descriptions to make them more aligned with tools' user scenarios, the agent bridges the gap between user queries and tool functionalities, enhancing the overall effectiveness of tool retrieval and usage.

In conclusion, this paper advances the field of tool integration with LLMs by introducing the novel Plan-and-Retrieve and Edit-and-Ground paradigms. Experiments show that our paradigms improve the recall and NDCG of tool retrieval tasks, significantly outperforming current state-of-the-art (SOTA). Our downstream evaluation suggests that the improvement gained during the retrieval phase, such as higher accuracy and relevance in responses, significantly contribute to successfully addressing the user queries.

\section{Related Works}
\stitle{Retrieval-Augmented LLM} Early studies on Retrieval-Augmented LLMs typically incorporate embeddings of retrieved passages 
as a part of the latent representation of the LM
\cite{chen2017reading, lee-etal-2019-latent}. More recent works like REALM \cite{guu2020retrieval}  and RAG \cite{lewis2021retrievalaugmented} have demonstrated the effectiveness of in-context augmentation and its improvement on knowledge-intensive tasks. There is also work \cite{mallen-etal-2023-trust} that explores how Chain-of-Thought (CoT) could guide a multi-turn Retrieve-and-Read process to solve open-domain questions and perform fact verification. 

However, the massive action space and tool functionality variance in tool-oriented tasks pose challenges to LLMs during planning. An erroneous step in planning can lead to a faulty loop, such as continually calling a tool in the wrong way or hallucinating non-existing tools. 
Our Plan-and-Retrieve paradigm, employing furtherest planning assessment \cite{zhu2023furthest}, enforces reasonable and goal-oriented decompositions of user queries. The recently proposed ReAct framework \cite{yao2022react} asks LLM to plan future actions based on its observation of environments. In the context of tool-oriented tasks, the plan builds upon the execution results of retrieved tools. Such practice running and verifying each tool at retrieval time can be expensive and time-consuming at scale. In contrast, our Plan-and-Retrieve paradigm fully leverages LLM's internal representation of world knowledge to propose plans in response to user queries, therefore guaranteeing both time and cost efficiency as an execution-free paradigm.


\stitle{Tool Learning}
Tool learning refers to the process where LLMs not only process and generate language-based responses but also learn to interact with and utilize external tools to enhance their capabilities \cite{nakano2022webgpt, schick2023toolformer, shen2023hugginggpt, qian2023creator, song2023restgpt, xu2023tool, li2023apibank, hao2023toolkengpt, zhang2023syntax}. By incorporating tools, LLMs can offer solutions in various areas, including visual-language processing \cite{gupta2023visual, wu2023visual}, mathematical reasoning \cite{lu2023chameleon}, and tasks in specialized domains \cite{jin2023genegpt,tang2023medagents}. 

However, previous research on tool learning mainly focused on teaching LLMs to use tools, but ignores the importance of shortlisting relevant tools. In this paper, we focus on using LLMs to improve the tool retrieval process.
In contrast to previous researches that heavily rely on finetuning retrievers \cite{schick2023toolformer, patil2023gorilla} to shortlist tools, we propose a novel Edit-and-Ground paradigm, leveraging LLMs' parametric knowledge to learn and create more informative descriptions for tools. 
This approach seeks to provide richer information for the retriever, leading to more accurate retrieval. 



\section{Task and Data}
We hereby formulate the task of tool retrieval and describe the dataset for this task.

\subsection{Task Definition}
The tool retrieval process involves taking a user query $Q$ and an index base of tool descriptions $D=\{d(t_1), d(t_2), \ldots, d(t_n)\}$ as input, where each $d(t)$ represents the description of each tool $t$. The retriever then sifts through the tool descriptions in $D$ and shortlists a relevant tool set $T = \{t_1, t_2, \ldots, t_k\}$ that are potentially suited to address aspects of the user query $Q$. It is essential to underline that unlike conventional retrieval tasks, the task of tool retrieval is goal-oriented in nature, which means the set of retrieved tools $T$ should be able to address the user query $Q$. 

The systems are expected to accurately retrieve relevant tools and understand the user intents and complex synergy between tools, thus truly assisting users in problem-solving processes.



\subsection{Dataset}
Existing datasets for tool learning, such as those delineated in \cite{li2023apibank, patil2023gorilla, tang2023toolalpaca, xu2023tool}, provide insights into the field. Nonetheless, these datasets exhibit limitations, where they only cover a limited number of tools or solely support simple single-tool usage scenarios, where user queries are simple and could be addressed by a single tool.

Contrastingly, \citet{qin2023toolllm} proposed ToolBench, a dataset covering more than 3,000 tools from 49 categories (such as advertising, data analysis, and transportation) and support complex, multi-tool user scenarios. In these scenarios, a single user query necessitates the sequential application of multiple tools, each contributing uniquely to the resolution of the query. The ToolBench dataset synergizes with the RapidAPI Hub, a prominent API marketplace that consolidates a vast array of real-world APIs. The multi-tool query creation process involves selecting representative tools within each category or collection, crafting queries to mimic real-world problem-solving scenarios.

Given our research focus and the nature of our study, we have chosen to concentrate on the \emph{Intra-Category} setting of the ToolBench dataset. The intra-category setting provides high-quality user queries, where the hierarchies of tools are clearly defined based on their main functionalities. It motivates understanding complex interactions and synergies between tools that share a common functional domain.
The setting mirrors real-world situations where problem-solving often demands a multifaceted and integrative use of diverse tools. The ToolBench dataset annotates paths of executed tools that successfully address the user queries as solution paths. The average length of the solution paths is 4. We take the annotated solution paths as the ground truth for our task.

\section{Method}
In this section, we describe the proposed framework to integrate tools with LLMs for addressing complex user queries. Our methodology is grounded in two innovative paradigms: the Plan-and-Retrieve (P\&R; \Cref{ssec:pr}) and Edit-and-Ground (E\&G; \Cref{ssec:eg}). We discuss the coordination between two paradigms in \Cref{ssec:inference}.

\subsection{Method Overview}\label{ssec:overview}
\modelname integrates two key paradigms, Plan-and-Retrieve (P\&R) and Edit-and-Ground (E\&G), to effectively address complex user queries with LLMs.

The \textbf{Plan-and-Retrieve} paradigm is a two-stage process. The \emph{Plan} stage decomposes user queries into focused sub-queries, while the \emph{Retrieve} stage matches these sub-queries with relevant tools. 

The \textbf{Edit-and-Ground} paradigm, consisting of the \emph{Evaluator} and \emph{Optimizer}, focuses on enhancing tool descriptions. 

These paradigms are designed to work in tandem. P\&R paradigm addresses immediate user queries, while E\&G actively identifies and collects under-informative tool descriptions for optimization.



\subsection{Plan-and-Retrieve}
\label{ssec:pr}
The Plan-and-Retrieve (P\&R) paradigm is designed as a two-stage process to effectively address complex user queries.

\stitle{Plan} In the \emph{Plan} stage, a LLM-based planner autoregressively decomposes the user query \( Q \) into sub-queries \( q_1, q_2, \ldots, q_n \). To ensure the robustness and quality of the decomposed sub-queries, we follow \citet{zhu2023furthest}. Specifically, for each step of sub-query generation, the planner first generates a batch of hypotheses. Then, we cluster the generated hypotheses along with previously created sub-queries via K-means clustering algorithm. Finally, we select a sub-query from the hypotheses that distinguishes the most from the previous sub-queries to proceed\footnote{Please refer to \Cref{ssec: kmeans} for algorithm implementation.}.

As shown in \Cref{fig:pipeline}, the planner autoregressively decomposes the user query \( Q \) into more fine-grained sub-queries based on assessments at inference time. 
After the generation of a sub-query \( q_{t} \), the planner evaluates whether the original query \( Q \) has been satisfactorily achieved based on the current planning history. If the evaluation determines that the goal has been met, the iterative process concludes. Otherwise, the planner proceeds to generate the subsequent sub-query \( q_{t+1} \). This active and autoregressive planning at inference time facilitates a more focused understanding of the tools. We use the following prompt template for the planner.

\begin{figure}[h]
    \centering
    \includegraphics[width=0.95\linewidth]{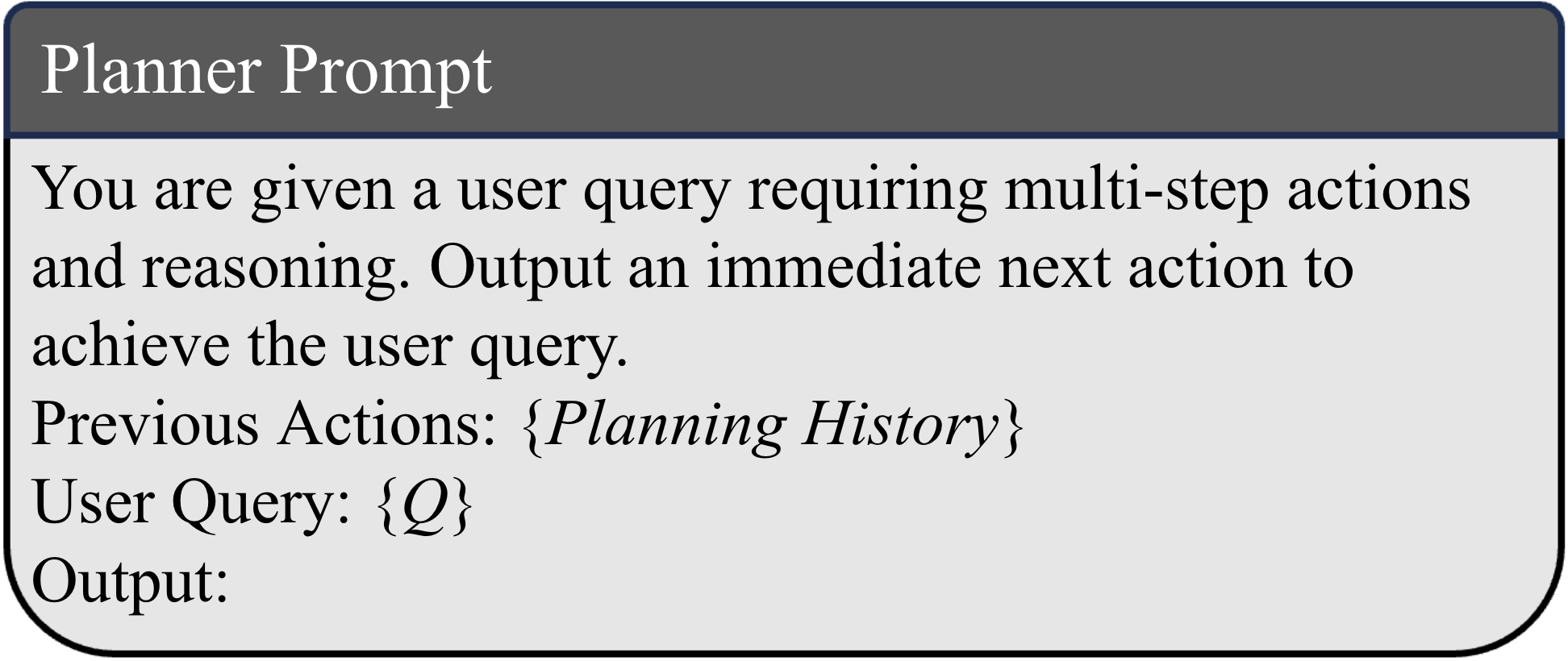}
    \label{fig:query_planning}
    \vspace{-1em}
\end{figure}

\stitle{Retrieve} In the \emph{Retrieve} stage, for each sub-query \( q_i \), the retriever shortlists the most suitable tools \( T_i \in D \). We first retrieve a pool of candidate tools that matches $q_i$, represented as 
\begin{equation}
     T'_i = Ret(q_i), 
\end{equation}
where $Ret$ represents the retriever.

To enhance the robustness of retrieval, we re-rank the candidate tool set $T'_i$ by LM-likelihood score between the sub-query $q_i$ and each tool $t_j \in T_i$, which is calculated as follows:
\begin{equation}
    \text{LM-likelihood}(q_i, t_j) = - \log P(q_i,d(t_j)).
\end{equation}
Based on the re-ranked tools, we choose the top-5 tools $T'_{i,top-5}$ and feed them into a LLM-based predictor, which outputs a shortlisted tool set $T_i$ from the candidate tool set $T'_{i,top-5}$ that are relevant to $q_i$. We use this prompt for the predictor. 

\begin{figure}[h]
    \centering
    \includegraphics[width=0.95\linewidth]{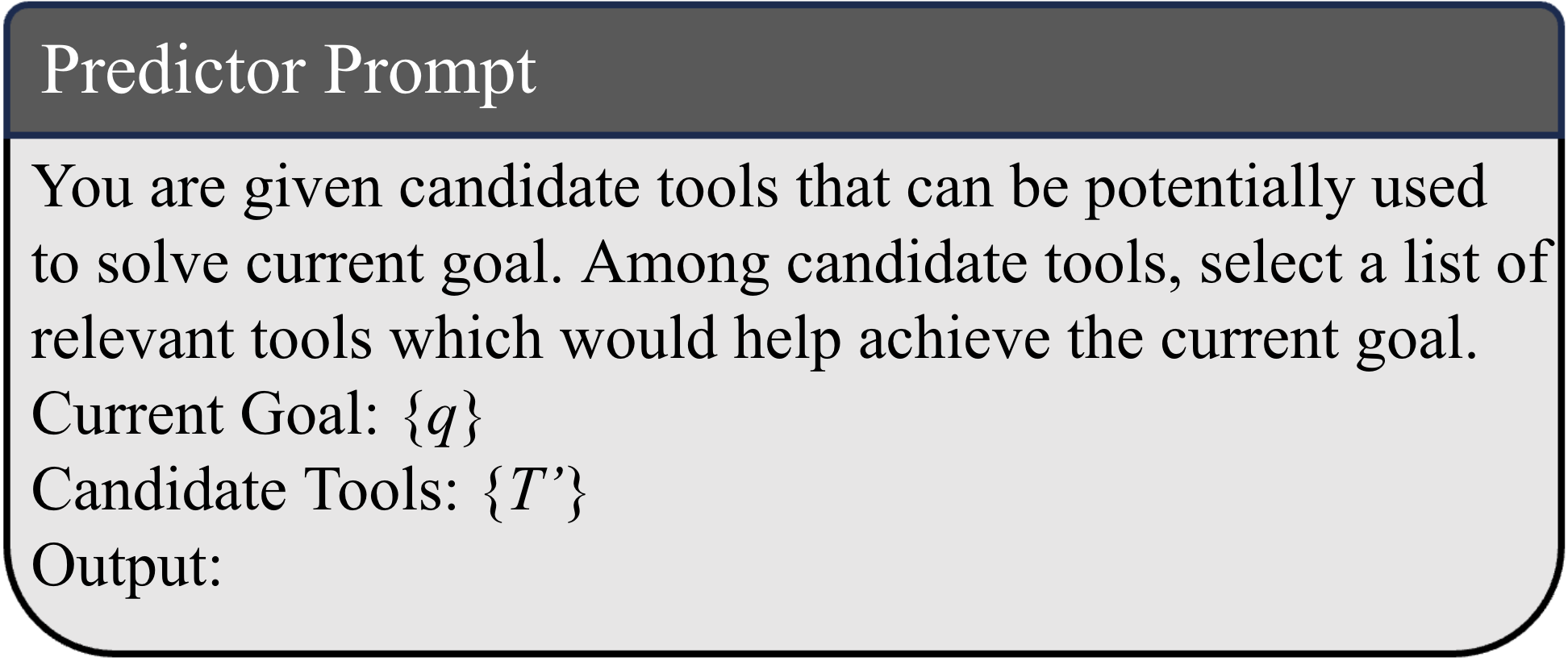}
    \label{fig:predictor_prompt}
    \vspace{-1em}
\end{figure}

As a result, the final shortlisted tool set $T$ is formed by \begin{equation}
     T = \bigcup_{i=1}^{n} T_i, \forall i \in [1,n] \cap \mathbb{Z}.
\end{equation}

For the choice of $Ret$, we adopt a neural (dense) retriever method. For each sub-query $q_i$, the dense vector representation $\mathbf{q}_i$ is obtained by passing $q_i$ through a dense encoder. Similarly, we obtain dense representation \textbf{d} through a dense encoder for each tool description $d$. The tool index corpus $D$ is formed as a collection of \textbf{d}.

The P\&R module interleaves \emph{Plan} and \emph{Retrieve} until the planner evaluates that the user query has been sufficiently decomposed and addressed through the retrieved tools. The module then returns $T$ as the relevant tools to address the user query.

\begin{algorithm}[t]
\caption{Edit-and-Ground Algorithm}
\begin{algorithmic}
\small
\State \textbf{Input:} Trainset, Devset, Toolset, Failure\_Threshold, Max\_Rounds
\State \textbf{Output:} Optimized Tool Descriptions

\State Initialize cache for tools in \textit{Toolset}
\State cur\_round = 0
\\
\While{cur\_round < Max\_Rounds}
    \State{\emph{\#\# Phase 1: Evaluate Retrieval Performance}}
    \For{each (\textit{query}, \textit{gt\_tools}) in \textit{Trainset}}
        \State \textit{predicted\_tools} $\leftarrow$ P\&R(\textit{query})
        \For{each \textit{tool} in \textit{gt\_tools}}
            \State \textit{tool.trials} += 1
            \If{\textit{tool} not in \textit{predicted\_tools}}
                \State \textit{tool.failure} += 1
                \State \textit{tool.queries}.add(query)\Comment{Failure queries}
            \EndIf
        \EndFor
    \EndFor
\\
\State{\emph{\#\# Phase 2: Failed Tool Description Optimization}}

    \For{each \textit{tool} in \textit{Toolset}}
        \If{$\frac{\textit{tool.failure}}{\textit{tool.trials}}$ > \textit{Failure\_Threshold}}
            \State \textit{$U$} $\leftarrow$ Remove specific entities from \textit{tool}.queries 
            \State \textit{$R$} $\leftarrow$ Predict \textit{reasons} for failure of $U$
            \State \textit{d(tool)} $\leftarrow$ \textit{tool.description}
            \State \textit{d'(tool)} $\leftarrow$ 
            E\&G(\textit{tool}, \textit{d(tool)}, U, R) \\
            \State{\emph{\#\# Phase 3: Evaluate Performance of d'(tool)}}
            \State \textit{cur\_recall} $\leftarrow$ Eval(\textit{Devset}, \textit{d'(tool)})
            \If{\textit{tool.recall} < \textit{cur\_recall}} 
                \State \textit{tool.description} $\leftarrow$ \textit{d'(tool)}
                \State \textit{tool.recall} $\leftarrow$ \textit{cur\_recall}
            \EndIf
        \EndIf
    \EndFor
    
    \State cur\_round += 1 
\EndWhile
\label{Algo: EG_selection}
\end{algorithmic}
\end{algorithm}

\subsection{Edit-and-Ground}
\label{ssec:eg}
The Edit-and-Ground (E\&G) paradigm focuses on refining under-informative tool descriptions to align them with user queries. As shown in \Cref{Algo: EG_selection}, the evaluator examines the quality of tool descriptions by retrieval results. A tool description is viewed as under-informative if the number of failure cases of retrieval exceeds a pre-defined threshold. We collect such tools for later optimization. 

\begin{figure}[h]
    \centering
    \includegraphics[width=0.95\linewidth]{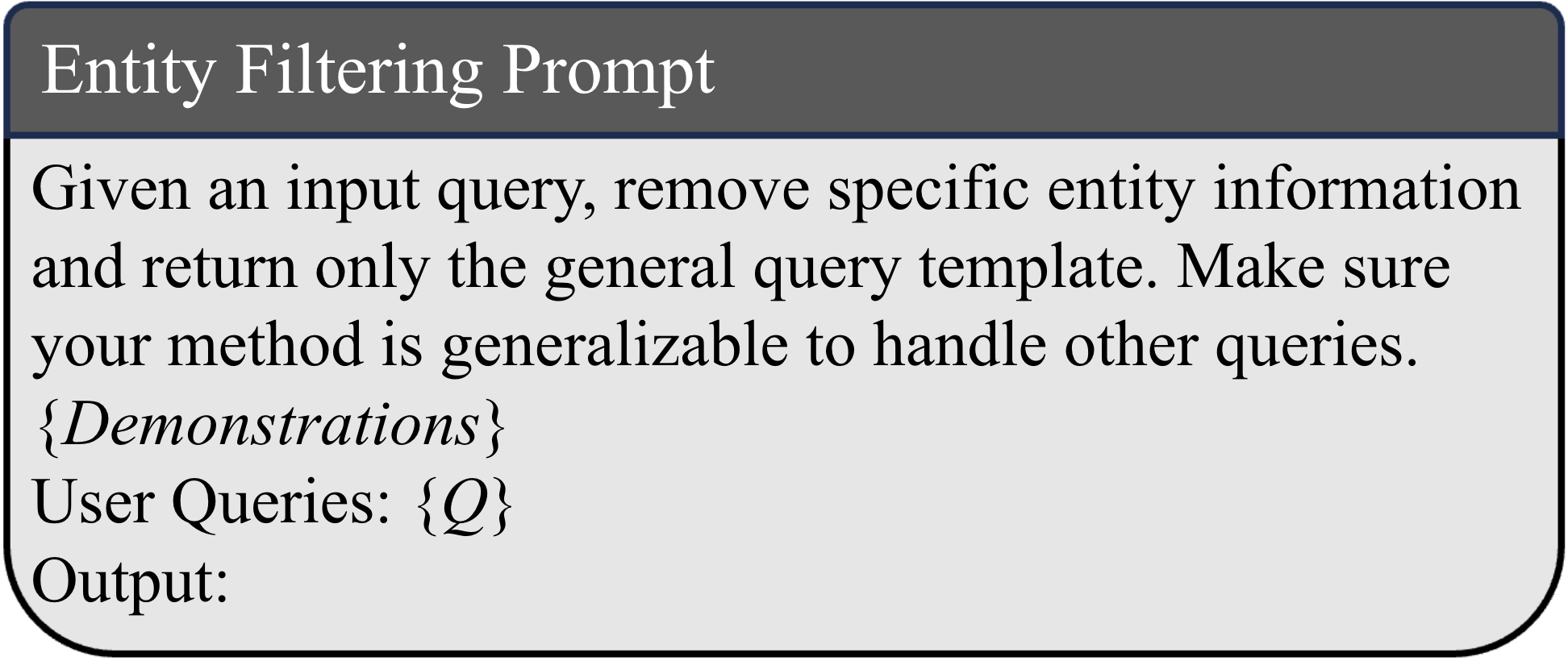}
    \label{fig:entity_filter}
    \vspace{-1em}
\end{figure}
Subsequently, the optimizer takes a tool \( t \) with its base description \( d(t) \) and $U$, a batch of relevant user queries, as input. To avoid the optimizer overfitting to a local batch, we use an LLM to filter out specific entities for each query in $U$. The entity filtering prompt template is shown as above.

To assist the optimizer in improving under-performed tool descriptions, we prompt LLM to generate reasons $R$ explaining why the tool could be related and helpful in addressing user queries. The functionality assessment prompt template is shown below:
\begin{figure}[h]
    \centering
    \includegraphics[width=0.95\linewidth]{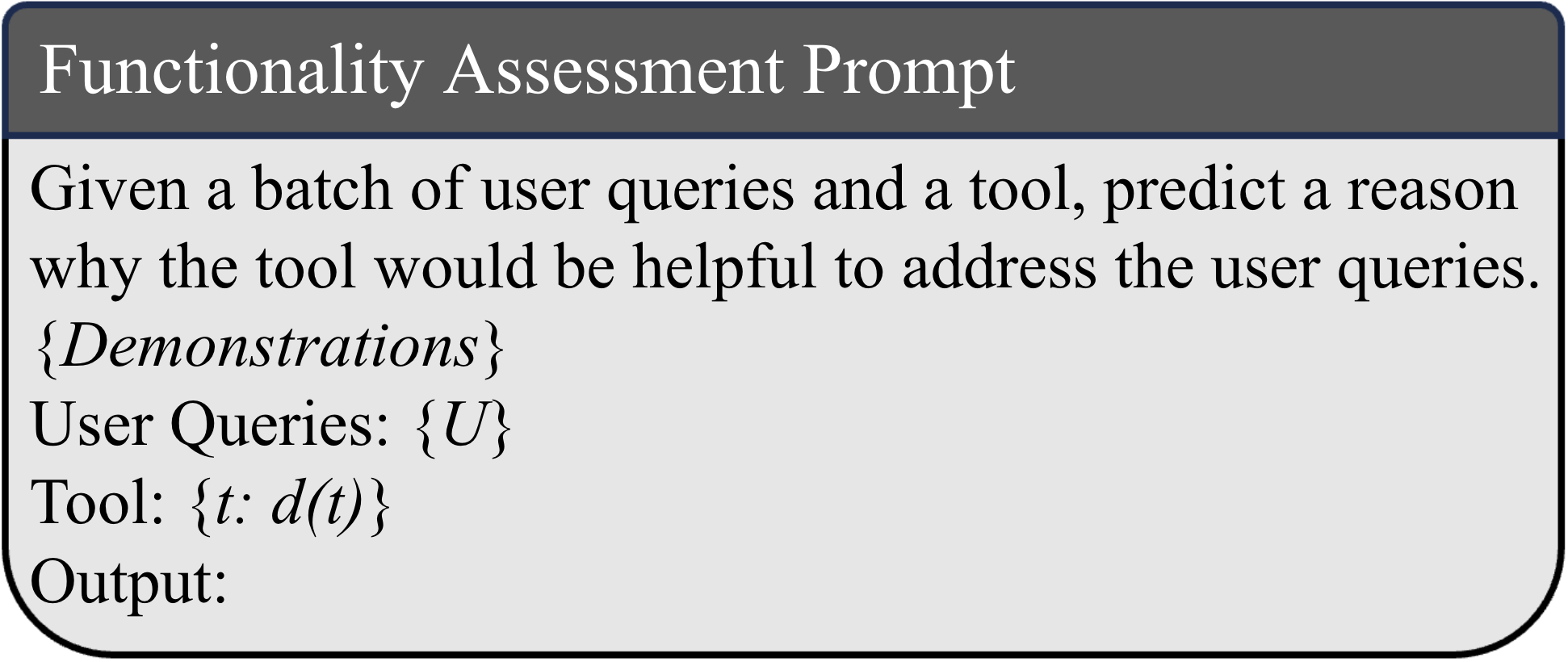}
    \label{fig:function_assess}
    \vspace{-1em}
\end{figure}

Finally, by prompting LLM with 1) base tool description $d(t)$, 2) entity-filtered user queries $U$, and 3) the reasons $R$, we obtain an enriched tool description \( d'(t) \). Please refer to \Cref{fig:edit-ground} in Appendix C for the prompt template. We formally represent this process as 
\begin{equation}
        d'(t) = E\&G(t, d(t), U, R).
\end{equation}

The optimization process is executed in multiple rounds as described in \Cref{Algo: EG_selection}. In each round, we evaluate the retrieval recall on the development set for each tool and compare it with the previous round. If the current round's recall is better than the previous one, we update the tool's description; otherwise, we keep the original description.

The Edit-and-Ground involves using the LLM's extensive world knowledge, combined with the contextual details provided by \( U \), to edit and enhance \( d(t) \). The result of this task is an enriched tool description \( d'(t) \), expected to resonate more closely with real-world user scenarios and increase the utility of the tool in practical applications.


\subsection{Paradigm Coordination and Inference}
\label{ssec:inference}

Our \modelname framework employs strategic coordination of the Plan-and-Retrieve (P\&R) and Edit-and-Ground (E\&G) paradigms, phased to optimize the process of tool retrieval. This section elucidates the interaction between these paradigms during the optimization phase and the subsequent inference phase.

\stitle{Optimization Phase}
During the optimization phase, P\&R and E\&G operate alternatively. P\&R is tasked with decomposing a user query \( Q \) into manageable sub-queries \( q_1, q_2, \ldots, q_n \). These sub-queries facilitate a more focused retrieval of tools from the tool set \( D \), ensuring that the process is aligned with specific aspects of the query.

During planning, the E\&G paradigm is actively engaged in optimizing the descriptions of the tools within \( D \). This optimization, leveraging the LLM's extensive knowledge base, is particularly targeted at tools that exhibit underperformance in retrieval effectiveness. By enriching these tool descriptions, E\&G significantly enhances the overall retrieval process, making the toolset more responsive and aligned with the practical demands of diverse queries.

\stitle{Inference Phase}
At the time of inference, the P\&R paradigm remains active, utilizing the previously enriched and optimized tool descriptions. 
In this phase, the E\&G paradigm ceases its
operation and does not engage in any further optimization of tool descriptions. 
The refined tool descriptions, already enhanced by E\&G, now serve as a comprehensive resource for the retriever to draw upon in response to the decomposed sub-queries.

\section{Experiments}
In this section, we evaluate the proposed \modelname framework for tool retrieval and compare it with baseline methods. We will delve into the details of our experimental setup (\Cref{ssec: exp_setup}), discuss the results (\Cref{ssec: main_result}) obtained, and perform an ablation study to understand strengths of different components (\Cref{ssec:ablation_study}). By executing the retrieved tools, we evaluate their correctness in addressing user queries to further validate our findings (\Cref{ssec: reader_exp}). We present case studies to qualitatively evaluate the strength of PLUTo framework (\Cref{ssec: case_study}).

\begin{table*}[t]
\centering
\setlength\tabcolsep{9pt}
\begin{tabular}{l|l|cccc}
\toprule
\multicolumn{1}{c|}{\multirow{2}{*}{\textbf{Model}}} & \multicolumn{1}{|c|}{\multirow{2}{*}{\textbf{Retriever}}} & \multicolumn{2}{c}{\textbf{Non-Finetuned}} & \multicolumn{2}{c}{\textbf{Finetuned}} \\
\cmidrule(lr){3-4} \cmidrule(lr){5-6}

 &  & \textbf{Rec} & \textbf{NDCG} & \textbf{Rec} & \textbf{NDCG} \\
\midrule
\multicolumn{1}{c|}{BM25} & \multicolumn{1}{|c|}{--} & 18.82 & 37.44 & -- & -- \\
\hline
\multicolumn{1}{c|}{\multirow{2}{*}{ToolRetriever}} & \multicolumn{1}{|c|}{DPR$^\dagger$} & 19.58 & 50.98 & 27.80 & 71.21 \\
& Contriever  & 31.78 & 74.70 & 42.77 & 79.16 \\
\hline
\multicolumn{1}{c|}{\multirow{2}{*}{\modelname}} & \multicolumn{1}{|c|}{DPR} &36.65 & 75.10 &  43.27 & 79.93\\
& Contriever & \textbf{46.57} & \textbf{82.93} & \textbf{48.47} & \textbf{84.73} \\
\bottomrule
\end{tabular}
\caption{This table compares various tool retrieval models using Recall and NDCG metrics in both Non-Finetuned and Finetuned settings. It includes an ablation study on the impact of using different retrievers, demonstrating the generalizability of \modelname. $\dagger$ indicates the previous SOTA implementation, as specified in \cite{qin2023toolllm}.}
\label{tbl:main}
\end{table*}

\subsection{Experiment Setup}
\label{ssec: exp_setup}
\stitle{Evaluation Protocol} We evaluate using three metrics to assess the effectiveness of our tool retrieval system. \emph{Recall (Rec)} measures the proportion of relevant tools that are successfully retrieved by our system. High indicates that the system is effective in identifying a comprehensive set of relevant tools for a given query and is more likely to yield a solution to address the user query.
We also report the \emph{Normalized Discounted Cumulative Gain (NDCG)} that evaluates the relevance and quality of ranked search results. In addition, we report \emph{pass rate}, an automatic evaluation metric of ToolBench \cite{qin2023toolllm}. The pass rate measures a system’s ability to successfully address the user query with a retrieved subset of tools in limited budgets by interacting with real-world RESTful APIs (\Cref{ssec: reader_exp}). 

To test the generalizability of our approach, we benchmark the tool retrieval performance under a Non-Finetuned setting, where we directly apply an off-the-shelf retriever model to comprehensively showcase \modelname's adaptivity. To test the model's practical applicability, we also benchmark retrieval performance under Finetuned setting, where we finetune the retriever model on domain-specific knowledge. We evaluate 500 user queries for each setting.

\stitle{Baselines} We compare our system against several representative retrieval methods. These include: (1) \emph{BM25:} a widely-used probabilistic retrieval framework, calculating the relevance of documents to a query based on the frequency of query terms in each document; 
(2) \emph{ToolRetriever}: a neural retrieval approach that 
achieves the current state-of-the-art (SOTA) performance on ToolBench retrieval task \cite{qin2023toolllm}.
To understand the flexibility of our framework, we benchmark \modelname's performance when incorporated with different retrievers. Specifically, we use DPR \cite{karpukhin-etal-2020-dense} and Contriever \cite{izacard2022unsupervised}.

\stitle{Implementation Details} 
For the implementation of PLUTo, we use DSPy framework \cite{khattab2023dspy} to facilitate efficient interaction between retriever and LLM. We choose ChatGPT\footnote{OpenAI. (2023). ChatGPT (November 21st version).} as our main LLM for both P\&R and E\&G. The maximum round for the E\&G module is set to 5. For ToolRetriever, we retrieve top-5 tools using the respective retrievers. The data is divided into 70-15-15 splits for training, development, and testing, respectively. For our experiment, we randomly select 500 data samples from the test split for each setting mentioned in Evaluation Protocol section.

For the Finetuned settings, we finetune the neural dense retriever model by including negative samples during in-batch training \cite{karpukhin-etal-2020-dense}. For each positive pair of query $q_j$ and its relevant tool $d_j^{+}$, we include $n$ negative tools as negative samples. We
use a cross-entropy loss with softmax function over the batch $B$:
\begin{equation}
L = 
- \frac{1}{B} \sum_{j=1}^{B} \log \left(\frac{e^{\mathbf{q}_j \cdot \mathbf{d}_j^{+}}}{e^{\mathbf{q}_j \cdot \mathbf{d}_j^{+}} + \sum_{i=1}^{n} e^{\mathbf{q}_j \cdot \mathbf{d}_{ij}^{-}}}\right)
\end{equation}

\subsection{Results}
\label{ssec: main_result}
The experimental results, detailed in \Cref{tbl:main}, underscore the significant advantages of our proposed \modelname models. In the Non-Finetuned setting, \modelname with Contriever showcases remarkable scores, achieving 46.57\% in Recall, outperforming the best baseline by 9.92 points. This result shows the model's robust ability to identify relevant tools without the necessity for specific finetuning, a critical advantage in dynamic tool retrieval environments. We observe a consistent trend in the Finetuned setting, with the model scoring 48.47\% in Recall, demonstrating a 5.7 points lead when compared with the Contriever baseline. This indicates that our model is highly effective on retrieving relevant tools. 

Furthermore, our model outperforms baselines across all settings on NDCG scores. In the Non-Finetuned setting, our model leads by 8.23 points. In the Finetuned setting, our model beats the baseline by 4.57 points. These results reflect \modelname not only the relevance of the tools retrieved but also their ranking in order of utility and applicability to the user's query, which is a indication to the model's nuanced understanding of tool utility.

To show the generalizability of \modelname, we select different retrievers for the Plan-and-Retrieve (P\&R) paradigm. We observe that \modelname has synergy with both DPR and Contriever models, regardless of their different architecture, that achieves higher Recall and NDCG scores than the baselines. This indicates that \modelname is a plug-n-play and retriever-agnostic framework that features effectiveness and flexibility under different circumstances.


The experimental results highlight the superior performance of \modelname framework. Together, the P\&R and E\&G paradigms establish a dynamic and effective framework, which not only accurately interprets and responds to user queries but also maintains an evolving understanding of tool functionality. This duality ensures that \modelname remains highly effective and adaptable in various setups, consistently aligning user needs with the most suitable tools and their capabilities.

\begin{figure}[t]
    \small
    \centering
    \includegraphics[width=1.02\linewidth]{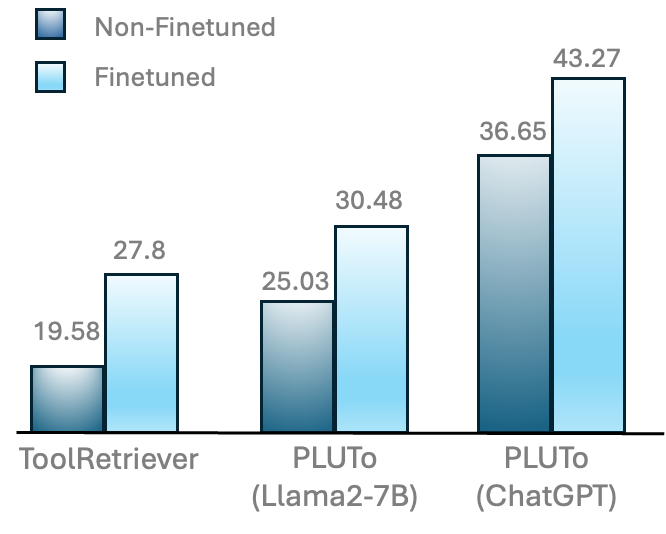}
    \caption{Performance comparison among different LLMs for Plan-and-Retrieve paradigm using Recall score. The backbone retriever is DPR.}     
    \label{fig:llama_ablation}
    \vspace{-1em}
\end{figure}

\subsection{Ablation Study}
\label{ssec:ablation_study}
As shown in \Cref{fig:llama_ablation}, we observe that both the Llama2 \cite{touvron2023llama} and ChatGPT variants show considerable improvements in tool retrieval capabilities, with notable increases in Recall and NDCG scores compared to baseline models. This consistent improvement across different LLM integrations conclusively demonstrates the robustness and effectiveness of our method. This finding is particularly important as it suggests that our approach is not overly reliant on any single LLM, thereby showcasing the broad applicability and potential of our methods in diverse settings.

As shown in \Cref{tbl:main_ablation}, the ablation experiment on the \modelname - full, focusing on the removal of Edit-and-Ground (E\&G) and Plan-and-Retrieve (P\&R) components, provides intriguing insights into their roles in tool retrieval tasks. Generally, removing E\&G leads to decreased Recall and NDCG scores across settings, underscoring its critical role in enhancing what the model seeks to retrieve.
On the other hand, excluding P\&R tends to diminish more of the model's performance in Non-Finetuned settings, particularly impacting Recall. This highlights P\&R's importance in effectively retrieving relevant information. A comparative analysis reveals that the full implementation of \modelname - ChatGPT, incorporating both E\&G and P\&R, consistently delivers strong performance across all metrics and settings, emphasizing the synergistic strength of these components. The variants of the model, lacking either E\&G or P\&R, provide valuable insights into the unique contributions of each component to the model's overall efficacy.
\begin{table}[t]

\centering
\setlength\tabcolsep{4.5pt}

\begin{tabular}{lcccc}
\toprule 
\multicolumn{1}{c|}{\multirow{2}{*}{\textbf{Model}}} & \multicolumn{2}{c}{\textbf{Non-Finetuned}} & \multicolumn{2}{c}{\textbf{Finetuned}} \\
\cmidrule(lr){2-3} \cmidrule(lr){4-5}

\multicolumn{1}{c|}{}& \textbf{Rec} & \textbf{NDCG} & \textbf{Rec} & \textbf{NDCG} \\
\midrule
\multicolumn{1}{c|}{\modelname - full} & \textbf{46.57} & \textbf{82.93} & \textbf{48.47} & \textbf{84.73} \\
\hline
\multicolumn{1}{c|}{- w/o E\&G} & 42.55 & 80.70 & 44.90 & 81.10 \\
\hline
\multicolumn{1}{c|}{- w/o P\&R} & 38.12 & 77.60 & 47.07 & 81.90 \\
\bottomrule
\end{tabular}
\caption{Ablation Study.}
\label{tbl:main_ablation}
\end{table}

\begin{table*}[]
\small
\centering
\begin{tabular}
{|m{0.3\linewidth}|m{0.2\linewidth}|m{0.2\linewidth}|m{0.2\linewidth}|}
\hline
 \centering \textbf{Question} & \centering  \textbf{Gold Answer} & \centering  \textbf{PLUTo Answer} & \textbf{ToolRetriever Answer}\\
\hline
I'm planning a weekend getaway with my partner and I want to surprise them with a romantic playlist. Could you fetch the reels and posts from romantic music artists on Instagram? Additionally, could you search for books about love and relationships on Open Library? & Instagram Reels and post Downloader, Open Library & \hl{Instagram Reels and post Downloader}, Instagram, \hl{Open Library}, Instagram Downloader & Love Quotes by LoveMelon, The Love Calculator, Book Finder, fb-video-reels, Reading Home APIs
\\
\hline
I'm planning a family movie night and I need a movie recommendation. Can you fetch the trending images for movie posters and provide me with the details of the most popular movie from the past month? Also, check the status of the movie session and download the completed movie. & Magisto, Bing Image Search & \hl{Magisto}, gogoanime-data-api, Youtube video info, Advanced Movie Search, Image Service, Memes, \hl{Bing Image Search}, Netflix Data & TikTok Info, Tiktok Video Feature Summary, TikTok Full Video Info, TikTok Downloader - Download Videos without watermark\\
\hline
I'm a music blogger and I'm searching for interesting radio stations to feature on my website. Can you help me find radio stations that play a mix of genres? Also, provide me with the details of the master for the track with the ID '987654' in the LANDR Mastering. & LANDR Mastering v1, 50K Radio Stations & 
GMC Radio, \hl{LANDR Mastering v1}, \hl{50K Radio Stations}, 60K Radio Stations &
\hl{LANDR Mastering v1}, Spotify\_v2, TuneIn, Spotify Scraper, Spotify\_v3\\
\hline
\end{tabular}
\caption{Performance comparison of PLUTo and ToolRetriever in retrieving relevant tools for user queries. This table demonstrates the effectiveness of PLUTo in closely aligning with the gold standard answers for diverse queries, showcasing its superior ability to understand and fulfill user needs compared to ToolRetriever. The highlighted tools are the correctly retrieved ones. 
}
\label{tab: case_study_table}
\end{table*}

\subsection{Execution Pass Rate}\label{ssec: reader_exp}

We evaluate the pass rate of the execution schema generated by ChatGPT using the DFSDT approach \cite{qin2023toolllm}. Using the ToolEval package, we assessed two distinct retrieval tools, ToolRetriever and \modelname, for their correctness and efficiency in responding to user queries. 
The \modelname achieves \textbf{72.3\%} for pass rate, while the previous SOTA system ToolRetriever scored \textbf{69.3\%}.

This experiment's findings emphasize the pivotal role of advanced retrieval strategies in enhancing user query response quality. The improvement gained during the retrieval phase, such as higher accuracy and relevance in responses, significantly contribute to the downstream tasks.

\subsection{Case Study}\label{ssec: case_study}
As shown in \Cref{tab: case_study_table}, we compare our \modelname against the ToolRetriever baseline to underscore \modelname's proficiency in retrieving relevant tools for diverse user queries. Through selected examples, \modelname's superior understanding and comprehensive response capabilities are highlighted, especially in scenarios requiring nuanced tool selection. 

For instance, for organizing a romantic weekend in the first example, \modelname not only identifies all essential tools but also enhances the search with additional relevant resources, showcasing its broad and accurate grasp of user needs. This is contrasted with ToolRetriever, where the retrieved tools are only similar on a surface level (the majority of the tools contain the term "Love") and fail to understand the user's intent. This emphasizes \modelname's improved relevance and precision in tool retrieval. We also showcase the descriptions of tools before and after optimization by the Edit-and-Ground paradigm in \Cref{tab: description_optimization_cases}.

By leveraging the Plan-and-Retrieve (P\&R) and Edit-and-Ground (E\&G) components, PLUTo marks a significant advancement over conventional retrieval systems, demonstrating its adaptability and utility in fulfilling diverse user requirements.

\section{Conclusion}
We introduced \modelname, a framework composed of the Plan-and-Retrieve and Edit-and-Ground paradigms, which marks a distinctive departure from traditional methodologies, setting a new standard for tool retrieval. The empirical results illustrate the superiority of \modelname across critical retrieval performance metrics 
as well as pass rate in real-world tool-use evaluation. These metrics collectively attest to the model's efficacy in identifying relevant tools and successfully addressing complex user queries. We hope the adaptability and efficiency of \modelname can empower a multitude of domains where accurate and timely retrieval of tools is paramount. From autonomous scientific discovery to software development, the potential applications are as diverse as they are impactful.

\section*{Acknowledgement}

We appreciate the reviewers for their insightful comments and suggestions.
Tenghao Huang and Muhao Chen were supported by an Amazon Research Award, a Keston Exploratory Research Award, and the NSF Grant ITE 2333736.

\section*{Limitation}
Our study, while enhancing tool learning by planning and editing strategies, is notably constrained by its reliance on English language datasets. This focus on English limits the model's applicability to other languages with distinct syntax and semantics and confines its evaluation to specific English data sources, leaving its performance on diverse language setups unexplored. Future research should address this limitation by developing multilingual capabilities and conducting evaluations across varied data sources. 

The Edit-and-Ground (E\&G) may be executed to further optimize the descriptions. However, due to the cost, we currently set a relatively loose stop criterion that is enough to demonstrate the effectiveness of the presented method.

\section*{Ethical Consideration}
In conducting this research, we have adhered to ethical guidelines and legal norms to ensure responsible data usage. The data used in this study was obtained from public datasets, specifically ToolBench. We ensured not to violate any terms of service of the data sources.

\bibliography{anthology,custom}
\bibliographystyle{acl_natbib}

\clearpage
\appendix
\section{K-means Algorithm for Furthest Planning}
\label{ssec: kmeans}
Here, we present the algorithm for selecting the optimal sub-query to proceed with the Plan-and-Retrieve paradigm.
\begin{algorithm}
\caption{Sub-query Selection}
\begin{algorithmic}
\State Let $Q_{\text{prev}}$ be the set of previous queries, and $Q_{\text{cand}}$ the set of candidate queries.
\If{$Q_{\text{prev}} = \emptyset$}
    \State \Return $Q_{\text{cand}}$
\EndIf
\State $Q_{\text{total}} = Q_{\text{cand}} \cup Q_{\text{prev}}$
\State $V = \text{TFIDFVectorizer}(Q_{\text{total}})$
\State $C = \text{KMeans}.\text{fit}(V)$
\State Let $L_{\text{prev}}$ be the cluster labels for $Q_{\text{prev}}$ in $C$.
\State $Q_{\text{filtered}} = \{ q \mid q \in Q_{\text{cand}}, \text{label}(q, C) \notin L_{\text{prev}} \}$
\If{$Q_{\text{filtered}} = \emptyset$}
    \State \Return $Q_{\text{cand}}$
\EndIf
\State \Return $\text{random.choice}(Q_{\text{filtered}})$
\end{algorithmic}
\end{algorithm}

\section{Evaluation Framework for NDCG Assessment}
In the process of evaluating the correspondence between the retrieved digital tools and the user's query, a nuanced approach is employed to assign relevance scores. This scoring paradigm operates on a scale from 0 to 2. A score of '2' is allocated exclusively to those tools that exhibit either an exact match or a functional equivalence to the predefined standards, referred to as 'ground-truths.' A score of '1' is designated for tools that are deemed to be of moderate relevance. Conversely, a score of '0' is reserved for tools that are determined to be irrelevant to the user's query. We hire graduate students to carry out this task.
\begin{figure*}[h]
    \small
    \centering
    \includegraphics[width=1.02\linewidth]{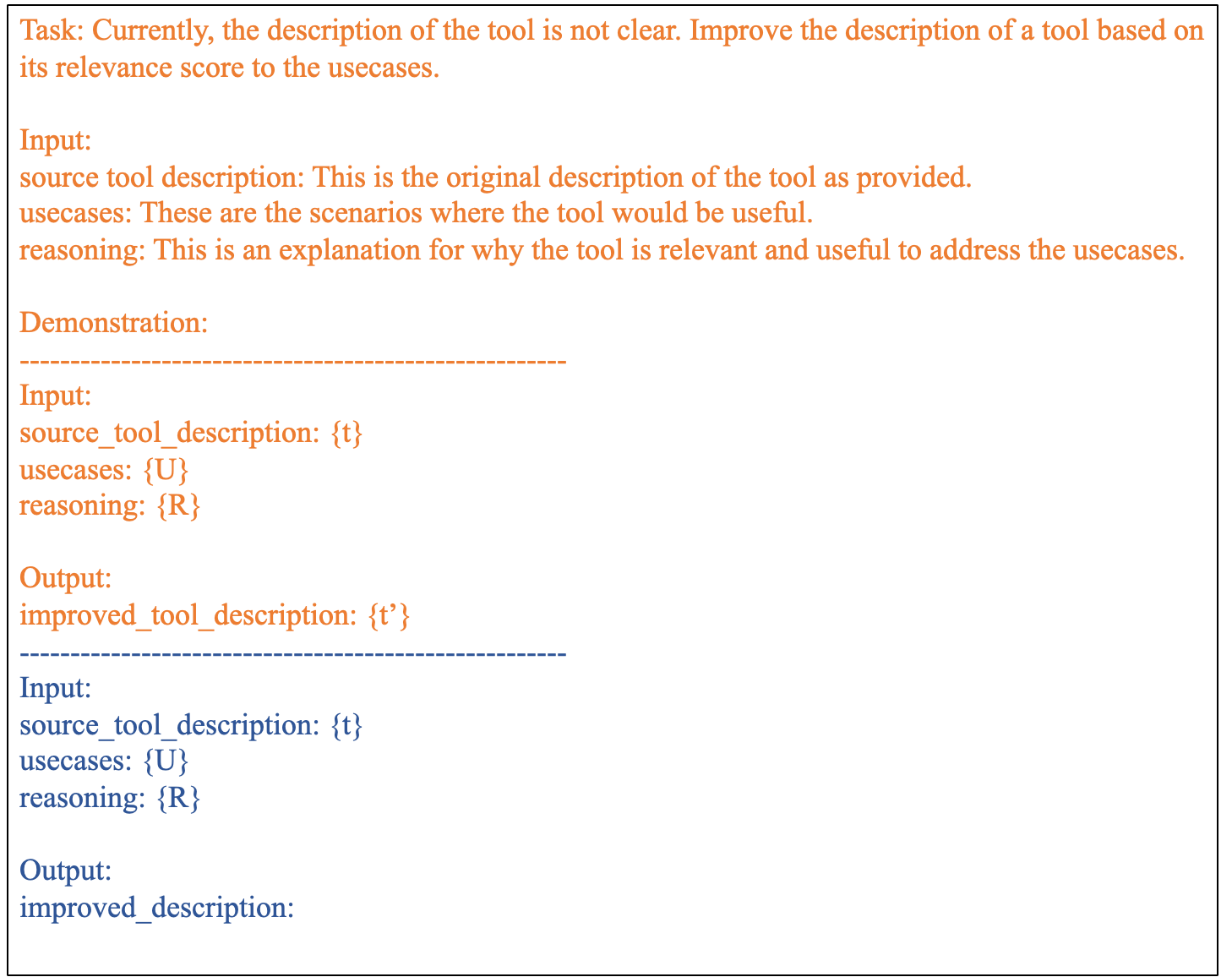}
    \caption{Edit-and-Ground template.}     
    \label{fig:edit-ground}
    \vspace{-1em}
\end{figure*}

\newpage
\section{Edit-Ground Prompt Template}
Please refer to \Cref{fig:edit-ground}. The task explanation and demonstration are shown in \textcolor{orange}{orange}. The input is shown in \textcolor{blue}{blue}.
\section{Case Studies for Edit-and-Ground paradigm}
From \Cref{tab: description_optimization_cases}, we observe that humans tend to craft tool descriptions that are abstract and simple, which can omit vital details necessary for both users and automated systems. Abstract descriptions can create a gap in understanding, particularly in domains where the nuances of functionality are key to effective use.

Contrastingly, tool descriptions that are contextualized and grounded by PLUTo offer a more detailed narrative, linking the tool's features with specific use cases and demonstrating its utility in practical scenarios. This detailed approach not only aids human users in understanding how a tool fits their specific needs but also enhances the precision of automated retrieval systems in matching tools to user queries. 

\begin{table*}[]
\small
\centering
\begin{tabular}
{|m{0.15\linewidth}|m{0.3\linewidth}|m{0.4\linewidth}|}
\hline
\textbf{Tool Name} & \textbf{Before} & \textbf{After}\\
\hline
Magisto &
Magically transform your videos.  Magisto turns your everyday videos into exciting, memorable movies you'll want to watch again and again. &
The Magisto API is a powerful tool that can transform user videos into exciting and memorable movies. It offers a range of capabilities to enhance the user's movie experience, provide movie recommendations, and optimize video content for discovery on platforms like Vimeo. The API allows users to choose visually appealing movies ...
\\
\hline
Bing Image Search &
An AI service from Microsoft Azure that turns any app into an image search resource. &
The Bing Image Search API is a comprehensive and powerful tool that enables users to search for and retrieve a wide range of images based on their specific requirements. With its extensive database and advanced search capabilities, this API offers a variety of features to cater to different use cases and scenarios. The API provides the ability to fetch trending images for movie posters. Additionally, the API offers the functionality to retrieve details of the most popular movie from the past month, enabling users to make informed decisions about their movie selection. 

\\
\hline
LANDR Mastering v1 &
Give your users an instant and customized audio mastering solution by harnessing the power of LANDR’s industry-leading, AI-driven mastering engine and its patented machine learning capabilities.  & 
The "LANDR Mastering v1" API is a comprehensive and powerful audio mastering solution designed to cater to the needs of music producers, music lovers, and anyone interested in high-quality audio production. With its instant and customizable features, the API offers a range of capabilities to enhance the music production process. For music producers, the API allows them to download a specific track by providing the track ID. Additionally, the API can fetch track credits, enabling the user to gather essential information for their music production needs ...

\\
\hline
\end{tabular}
\caption{Comparative Analysis of Tool Descriptions before and after optimization by the Edit-and-Ground paradigm. This table showcases the evolution of tool descriptions for selected tools, illustrating the substantial enhancement in clarity, detail, and functionality offered to users.
}
\label{tab: description_optimization_cases}
\end{table*}

\end{document}